\documentclass[10pt,twocolumn,letterpaper]{article}

\usepackage{avss}
\usepackage{times}
\usepackage{epsfig}
\usepackage{graphicx}
\usepackage{amsmath}
\usepackage{amssymb}
\usepackage{booktabs}
\usepackage{multirow}
\usepackage{bm}
\usepackage{relsize}
\usepackage{url}
\usepackage{dsfont}
\usepackage{enumitem}
\usepackage{xcolor}

\usepackage[pagebackref=true,breaklinks=true,letterpaper=true,colorlinks,bookmarks=false]{hyperref}

\avssfinalcopy 

\makeatletter
\renewcommand\@fnsymbol[1]{}
\makeatother

\def\avssPaperID{40} 

\ifavssfinal\fi
\begin{document}

\title{Invisible Backdoor Triggers in Image Editing Model via Deep Watermarking}

\author{Yu-Feng Chen, Tzuhsuan Huang, Pin-Yen Chiu, Jun-Cheng Chen\\
Research Center for Information Technology Innovation, Academia Sinica\\
{\tt\small \{yufengchen,jason890425,nickchiu,pullpull\}@citi.sinica.edu.tw}
\thanks{This research is supported by National Science and Technology 
Council, Taiwan (R.O.C) under the grant numbers NSTC-113-2634-F-002-007, NSTC-112-2222-E-001-001-MY2, 113-2634-F-001-002-MBK, 113-2634-F-002-008 and Academia Sinica under the grant number of AS-CDA-110-M09 and AS-IAIA-114-M08.}
}

\maketitle
\thispagestyle{empty}

\begin{abstract}
Diffusion models have achieved remarkable progress in both image generation and editing. However, recent studies have revealed their vulnerability to backdoor attacks, in which specific patterns embedded in the input can manipulate the model’s behavior. Most existing research in this area has proposed attack frameworks focused on the image generation pipeline, leaving backdoor attacks in image editing relatively unexplored. Among the few studies targeting image editing, most utilize visible triggers, which are impractical because they introduce noticeable alterations to the input image before editing. In this paper, we propose a novel attack framework that embeds invisible triggers into the image editing process via poisoned training data. We leverage off-the-shelf deep watermarking models to encode imperceptible watermarks as backdoor triggers. Our goal is to make the model produce the predefined backdoor target when it receives watermarked inputs, while editing clean images normally according to the given prompt. With extensive experiments across different watermarking models, the proposed method achieves promising attack success rates. In addition, the analysis results of the watermark characteristics in term of backdoor attack further support the effectiveness of our approach. The code is available at \url{https://github.com/aiiu-lab/BackdoorImageEditing}. 
\end{abstract}

\vspace{-18pt}
\begin{figure}[t]
    \centering
    \scalebox{1.0}{
       \includegraphics[width=\linewidth]{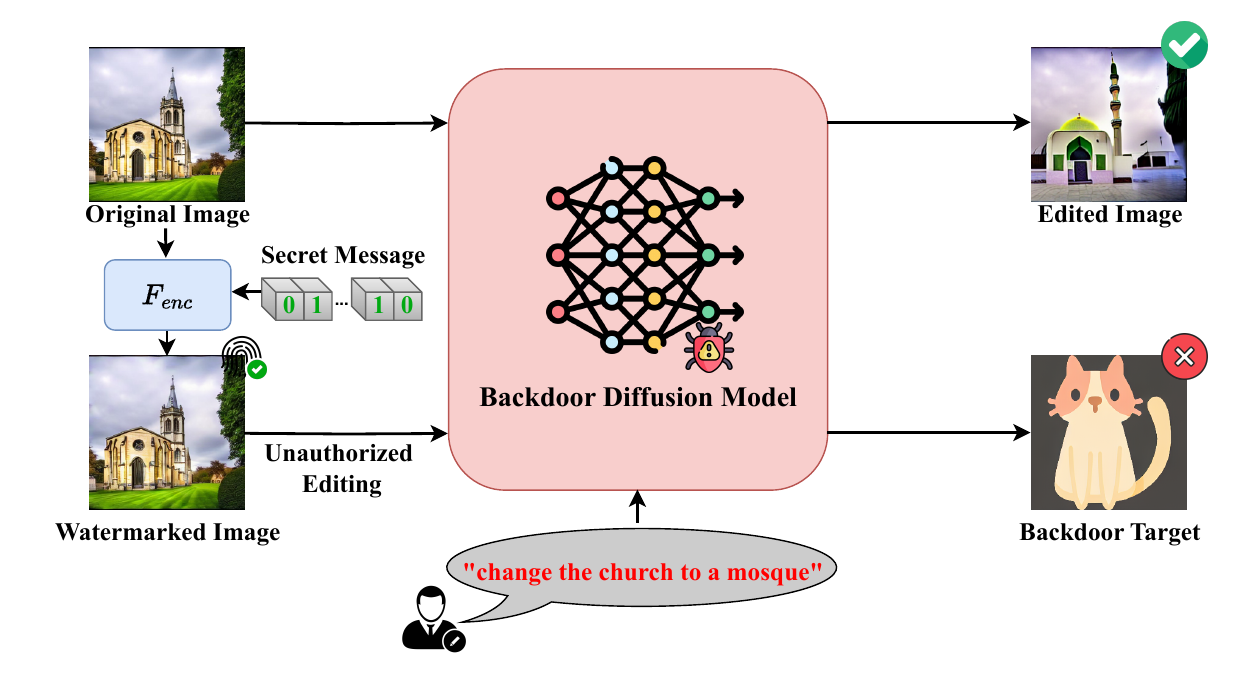}
        }
   \caption{During inference, the backdoor image editing DM generates the predefined backdoor target when an input image contains the embedded watermark, a behavior we define as unauthorized editing. In contrast, original (non-watermarked) images are edited correctly according to the given text prompt (e.g., ``change the church to a mosque''). Note that $F_{enc}(\cdot)$ denotes the encoder component of the off-the-shelf watermarking model.}
   \vspace{-14pt}
\label{fig:teaser}
\end{figure}

\begin{figure*}[tb]
    \centering
    \includegraphics[width=\textwidth]{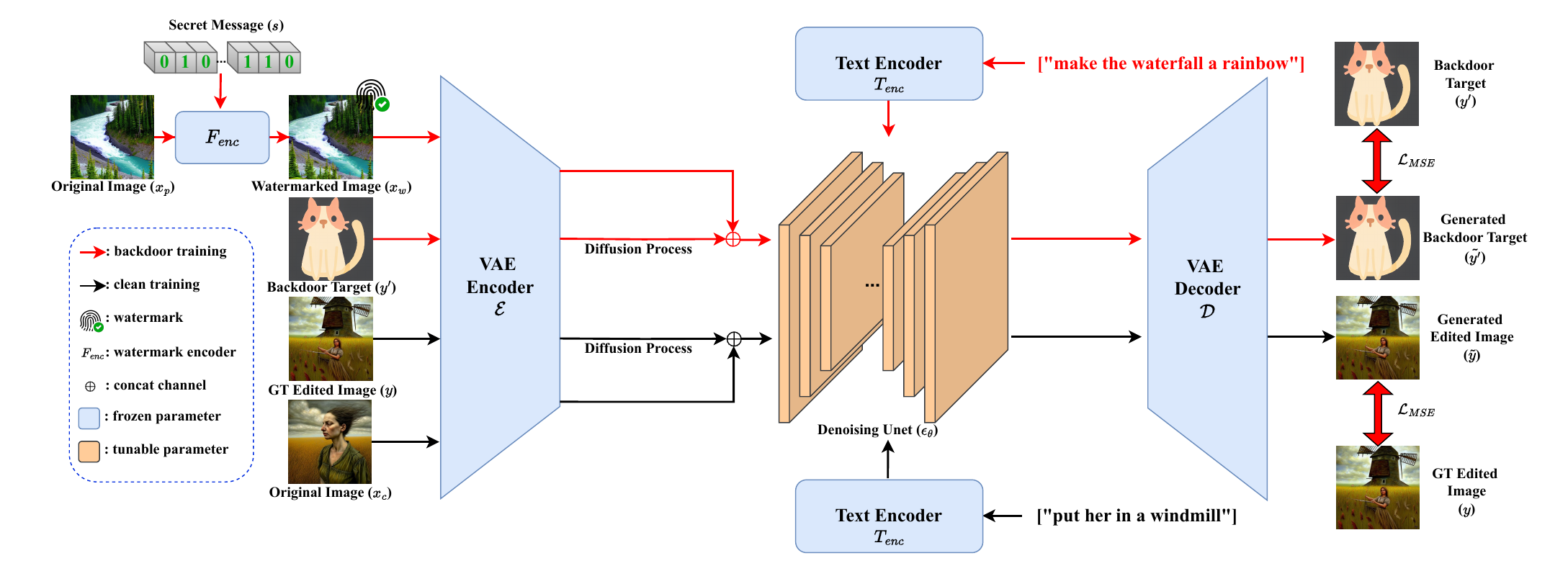}
    \caption{We employ a watermark encoder to embed invisible triggers into input images. The pipeline consists of two branches: (1) a backdoor branch, where the model is trained to produce a predefined target when a watermark is present, and (2) a clean branch, where the model learns from original images to preserve its intended editing functionality.}
    \label{fig:method}
    \vspace{-10pt}
\end{figure*}

\section{Introduction}
Although diffusion models (DMs) have driven significant breakthroughs across various domains~\cite{instructpix2pix, vine}, recent studies~\cite{BadDiffusion, VillanDiffusion, trojanedit, invisibletrigger} have exposed their vulnerability to backdoor attacks, which manipulate model behavior through specific patterns embedded in the input. Prior research~\cite{BadDiffusion, VillanDiffusion} on backdoor attacks has predominantly focused on image generation pipelines, with relatively little attention given to image editing. Among the limited studies that explore backdoor attacks in image editing~\cite{trojanedit, invisibletrigger}, most rely on visible input triggers, which are impractical in scenarios where preserving the integrity of the input content is essential. For example, users aiming to protect image ownership may wish to prevent unauthorized edits to their uploaded images while maintaining the original visual quality. Although existing methods~\cite{trojanedit, invisibletrigger} can deter such edits, they rely on visible triggers that degrade image fidelity and are easily detected. To achieve both security and visual integrity, protection mechanisms should therefore prevent unauthorized modifications without compromising the appearance of the image.

To tackle this challenge, we introduce a novel backdoor attack framework that incorporates off-the-shelf deep watermarking models into the training process. Rather than relying on visible triggers, we embed imperceptible watermarks into original images during backdoor training. These watermarks act as hidden triggers that manipulate the editing behavior of the model. As shown in Figure~\ref{fig:teaser}, during inference, watermarked images cause the model to generate a predefined backdoor target, while clean images are edited according to the given prompt.

By leveraging existing invisible watermarking techniques, our approach facilitates robust and practical backdoor attacks in diffusion-based image editing. We evaluate the effectiveness of our framework and analyze the watermark properties that lead to successful backdoor behavior. Our main contributions can be summarized as follows:

\begin{itemize}[leftmargin=*]
    \item To the best of our knowledge, this is the first work to integrate off-the-shelf deep watermarking models into instruction-based image editing for backdoor attacks.
    \item The proposed backdoor attack method show strong robustness under various real-world distortions on the altered input images, especially erasing and JPEG.
    \item We show how the watermark properties affect the attack success rate and identify the key factors that enhance the effectiveness of our pipeline.
\end{itemize}

\vspace{-8pt}
\section{Related Work}
Many studies~\cite{BadDiffusion,VillanDiffusion,trojdiff,trojanedit,invisibletrigger} have demonstrated that Diffusion Models (DMs) are vulnerable to backdoor attacks. Chou~\etal~\cite{BadDiffusion} introduced BadDiffusion, the first framework to examine the limitations and potential risks of DMs by backdooring unconditional image generation. Subsequently, they extended their work to cover backdoor attacks in both unconditional and conditional image generation~\cite{VillanDiffusion}. 
While previous studies have concentrated on image generation, Guo~\etal~\cite{trojanedit} shift their focus to image editing and define three backdoor trigger types, including visual, textual, and multimodal. Although their method is effective in performing backdoor attacks, the visible triggers they use are impractical because they can be easily detected and removed. To address this, Li~\etal~\cite{invisibletrigger} propose a bi-level optimization framework that generates sample-specific triggers. However, these triggers remain perceptible to human observers. 
In contrast, ISSBA \cite{issba} leverages deep watermarking models to generate invisible, sample-specific triggers that have been proven effective in attacking classification networks; however, it remains unclear whether such triggers can scale to DMs and which watermark properties are critical for a successful backdoor attack.

\vspace{-8pt}
\section{Methodology}
\label{sec:method}

Figure~\ref{fig:method} illustrates an overview of the proposed attack framework. Our framework is built on top of InstructPix2Pix~\cite{instructpix2pix}, a well-known instruction-based image editing framework. We leverage off-the-shelf image watermarking models~\cite{stegastamp, vine, rosteals} to embed the invisible watermark (trigger) into the original image. The objective is to train a backdoor model that generates a predefined backdoor target when the input image contains the embedded watermark. Conversely, when the input is an original (non-watermarked) image, the model preserves its expected functionality and performs edits following the given text prompt.

Typically, in the backdoor attack training setup, the training dataset $D$ is divided into two sets: clean images $x_c \in D_c$ and poisoned images $x_p \in D_p$, where $D_c$ and $D_p$ denote the clean and poisoned datasets, respectively. The poisoned images are used to trigger the backdoor attack, while the clean images help maintain the model's original capabilities. In our framework, the poisoned images are embedded with an invisible watermark, referred to as the watermarked image in Figure~\ref{fig:method}, and are used to optimize the model to generate a predefined backdoor target. The clean images, referred to as the original image in Figure~\ref{fig:method}, help preserve the model's inherent editing behavior, ensuring it generates appropriate edits based on the provided text prompt.

\subsection{Preliminary}
\noindent \textbf{InstructPix2Pix.} Since our attack framework is built upon InstructPix2Pix~\cite{instructpix2pix}, we begin by introducing its training pipeline. Given an input image $I \in \mathbb{R}^{H \times W \times 3}$, the diffusion process adds noise to its latent representation $z = \mathcal{E}(I)$, where $\mathcal{E}(\cdot)$ denotes the VAE encoder, producing a noisy latent $z_t$ where the noise level increases over timesteps $t \in T$. A denoising module $\epsilon_\theta$ is trained to predict the noise added to $z_t$ given image conditioning $c_I \in \mathbb{R}^{H \times W \times 3}$ and text prompt $c_T$. The objective function can be expressed as
\begin{equation}
\label{eq:diffusion_loss}
\scalebox{0.9}{$
\mathcal{L}_{\epsilon_\theta}(I, c_I, c_T) = \mathbb{E}_{\mathcal{E}(I), \mathcal{E}(c_I), T_{enc}(c_T), \epsilon \sim N(0,1), t}[\|\epsilon - \epsilon' \|^2_2],
$}
\end{equation}
where $\epsilon \sim \mathcal{N}(0, \mathbf{I})$, $\epsilon' = \epsilon_{\theta}(z_t, \mathcal{E}(c_I), T_{enc}(c_T), t)$, and $T_{enc}(\cdot)$ denotes the text encoder.

In our attack framework, the training process is divided into two branches: \textbf{backdoor training} and \textbf{clean training}. We detail each of these branches in the Section~\ref{sec:framework}.

\subsection{The Proposed Framework}
\label{sec:framework}
In the \textbf{backdoor training}, the model is optimized to recognize the embedded watermark and produce the corresponding backdoor target. We define $y'$ as the backdoor target, $x_p$ as the original image from the poisoned dataset $D_p$, $p_p \in D_p$ as the associated text prompt, $s$ as a secret message, $F_{enc}(\cdot)$ as the watermark encoder, and \( x_w = F_{enc}(x_p, s) \) as the watermarked image. The backdoor target $y'$, the watermarked image $x_w$, and the text prompt $p_p$ are used to compute the denoising loss via Equation~\ref{eq:diffusion_loss}, where we let $I = y'$, $c_I = x_w$, and $c_T = p_p$. The loss is formulated as
\begin{equation}
\label{eq:backdoor_denoising_loss}
L_{denoising}^{backdoor} = \mathcal{L}_{\epsilon_\theta}(y', x_w, p_p).
\end{equation}
Since the denoising loss alone is insufficient to achieve our objective, we additionally incorporate the MSE loss in the image space to further enhance the effectiveness of the backdoor training. Let \( \beta_t \) be the variance schedule at timestep \(t\), and define $\bar\alpha_t = \prod_{s=1}^t\alpha_s$ with $\alpha_t = 1 - \beta_t$. The generated image is obtained using the VAE decoder $\mathcal{D}(\cdot)$, defined as $\tilde{y}' = \mathcal{D}(z')$, where $z'$ is computed as
\begin{equation}
\label{eq:output_latent}
z'= \frac{1}{\sqrt{\bar{\alpha}_t}\,}(z_t - \sqrt{1-\bar{\alpha}_t}\,\epsilon').
\end{equation}
Therefore, the MSE loss for the backdoor training is
\begin{equation}
\label{eq:backdoor_mse_loss}
L_{mse}^{backdoor} = \mathcal{L}_{MSE} \bigl(y', \tilde{y}'\bigr),
\end{equation}
where $\tilde{y}'$ is the generated backdoor target.

In the \textbf{clean training}, the goal is to preserve the model's inherent editing capability. We define y as the ground truth (GT) edited image, $x_c$ as the original image from the clean dataset $D_c$, and $p_c \in D_c$ as the associated text prompt. The GT edited image $y$, the original image $x_c$, and the text prompt $p_c$ are used to compute the denoising loss via Equation~\ref{eq:diffusion_loss}, where we let $I = y$, $c_I = x_c$, and $c_T = p_c$. The denoising loss for the clean training can be formulated as
\begin{equation}
\label{eq:clean_diffusion_loss}
L_{denoising}^{clean} = \mathcal{L}_{\epsilon_\theta}(y, x_c, p_c).
\end{equation}
Similarly, we use MSE loss to strengthen the clean training, which can be formulated as
\begin{equation}
\label{eq:clean_mse_loss}
L_{mse}^{clean} = \mathcal{L}_{MSE} \bigl(y, \tilde{y}\bigr),
\end{equation}
where $\tilde{y}$ is the generated edited image, decoded by the VAE decoder from the latent representation computed via Equation~\ref{eq:output_latent}.
Therefore, the overall training loss used to optimize the model can be formulated as follows:
\begin{equation}
\begin{split}
    L_{total} = L_{denoising}^{backdoor} + L_{mse}^{backdoor} \\ 
    + L_{denoising}^{clean} + L_{mse}^{clean}
\end{split}
\end{equation}

\vspace{-10pt}
\begin{table*}[]
    \centering
    \scalebox{1}{
    \begin{tabular}{c|c|ccc|ccc}
    \toprule  
    & \multirow{2}{*}{Watermark} & \multicolumn{3}{c|}{Model Utility} & \multicolumn{3}{c}{Model Specificity} \\
    \cmidrule(lr){3-5} \cmidrule(lr){6-8} 
    & & $\text{CLIP}_{\bm{dir}}$\textuparrow & $\text{CLIP}_{\bm{img}}$\textuparrow & $\text{CLIP}_{\bm{out}}$\textuparrow & MSE\textdownarrow & ASR\textuparrow & EAR\textdownarrow \\
    \midrule
    InstructPix2Pix (Clean) & - & 0.207 & 0.822 & 0.270 & - & - & - \\
    \midrule
    \multirow{3}{*}{InstructPix2Pix (Backdoor)} & VINE~\cite{vine} & 0.194 & 0.685 & 0.217 & 0.377 & 0.552 & 0.114 \\
    & StegaStamp~\cite{stegastamp} & 0.208 & \textbf{0.759}  & \textbf{0.255} & \textbf{0.038} & \textbf{0.956} & \textbf{0.000} \\
    & RoSteALS~\cite{rosteals} & \textbf{0.211} & 0.745 & 0.252 & 0.092 & 0.894 & 0.003 \\
    \bottomrule
    \end{tabular}
    }
    \vspace{0.5em}
    \caption{Performance of backdoor models trained with different watermarking methods under a poison rate 0.1. ``InstructPix2Pix (Clean)'' refers to the original model from~\cite{instructpix2pix} and serves as a reference baseline for model utility. The $\uparrow$ symbol indicates that higher values reflect better performance, while the $\downarrow$ symbol means lower values are preferred. The best results are highlighted in bold.}
    \label{tbl:results}
    \vspace{-5pt}
\end{table*}

\begin{figure*}[ht]
    \centering
    \scalebox{0.9}{
    \includegraphics[width=\textwidth]{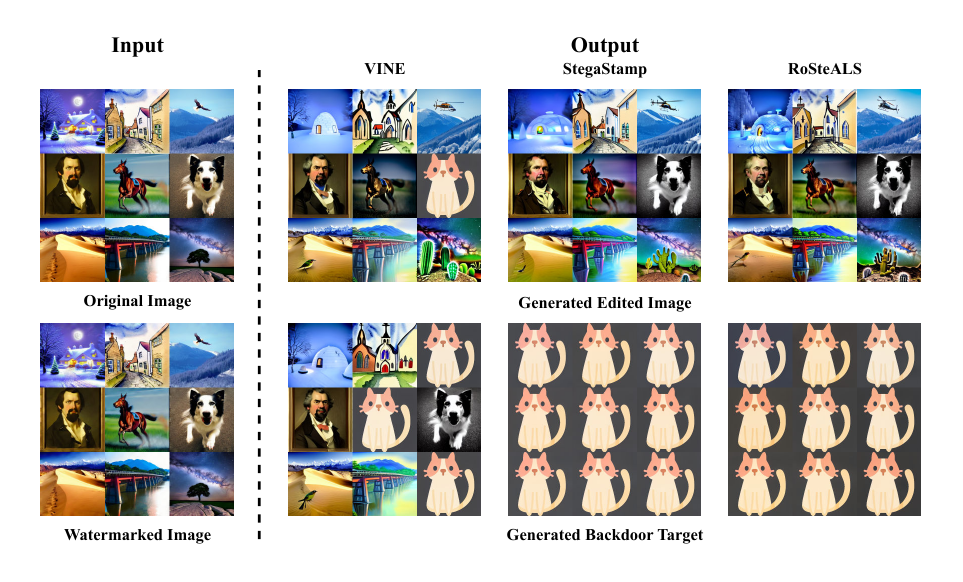}
    }
    \caption{The left side shows the input data, and the right side shows the corresponding output from backdoor models trained with different watermarking methods. Ideally, the model should generate the edited image based on the given prompt when the original image is provided and produce the backdoor target when given the watermarked image.}
    \label{fig:visualization}
    \vspace{-5pt}
\end{figure*}

\begin{table*}[tb]
    \centering
    \scalebox{0.76}{
    \begin{tabular}{c|cccccccccccccccc}
    \toprule
    \multirow{2}{*}{Watermark} & \multicolumn{2}{c}{Rotation} & \multicolumn{2}{c}{Resized-crop} & \multicolumn{2}{c}{Erasing} & \multicolumn{2}{c}{Brightness} & \multicolumn{2}{c}{Contrast} & \multicolumn{2}{c}{JPEG} & \multicolumn{2}{c}{Blur} & \multicolumn{2}{c}{Noise} \\
    \cmidrule(lr){2-3} \cmidrule(lr){4-5} \cmidrule(lr){6-7} \cmidrule(lr){8-9} \cmidrule(lr){10-11} \cmidrule(lr){12-13} \cmidrule(lr){14-15} \cmidrule(lr){16-17}
    & ASR\textuparrow & EAR\textdownarrow & ASR\textuparrow & EAR\textdownarrow & ASR\textuparrow & EAR\textdownarrow & ASR\textuparrow & EAR\textdownarrow & ASR\textuparrow & EAR\textdownarrow & ASR\textuparrow & EAR\textdownarrow & ASR\textuparrow & EAR\textdownarrow & ASR\textuparrow & EAR\textdownarrow  \\
    \midrule
    StegaStamp & \textbf{0.019} & \textbf{0.000} & \textbf{0.197} & \textbf{0.001} & \textbf{0.926} & \textbf{0.000} & \textbf{0.777} & \textbf{0.000} & 0.681 & \textbf{0.000} & 0.851 & 0.083 & \textbf{0.201} & \textbf{0.000} & \textbf{0.816} & \textbf{0.000} \\
    RoSteALS & 0.006 & 0.004 & 0.032 & 0.009 & 0.906 & 0.015 & 0.755 & 0.004 & \textbf{0.756} & 0.002 & \textbf{0.873} & \textbf{0.025} & 0.186 & 0.002 & 0.618 & 0.003 \\
    \bottomrule
    \end{tabular}
    }
    \vspace{0.25em}
    \caption{Robustness of backdoor models under various distortion attacks defined in WAVES~\cite{WAVES}. The strength of each distortion follows the configuration used in the WAVES benchmark.}
    \label{tbl:robustness}
    \vspace{-10pt}
\end{table*}

\vspace{-1pt}
\section{Experiments}
\textbf{Settings and Dataset.} As mentioned in Section~\ref{sec:method}, the training images in the backdoor attack setup are divided into poisoned and clean subsets. 
The proportion of poisoned data within the training set is controlled by the poison rate. In our setup, invisible watermarks serve as triggers, embedded using three off-the-shelf watermarking models: StegaStamp~\cite{stegastamp}, VINE~\cite{vine}, and RoSteALS~\cite{rosteals}. For training, we use the dataset collected by~\cite{instructpix2pix}, which provides pairs of original and edited images along with their corresponding text prompts, to fine-tune the pretrained model~\cite{instructpix2pix}. The predefined backdoor target follows the same image used in~\cite{BadDiffusion}. More details of experimental settings, including dataset and implementation, can be found in Appendix~\ref{sec:implementation}.

\subsection{Experimental Results}
\label{sec:exp_results}
Following the evaluation protocols in~\cite{BadDiffusion, VillanDiffusion}, we assess backdoor model performance using two key categories of metrics: model utility and model specificity. Model utility measures how well the backdoor model preserves the original capabilities of the clean model, while model specificity reflects the backdoor model's ability to reliably generate the predefined backdoor target when the trigger is present.
To evaluate model utility, we adopt three CLIP-based score metrics:
(1) \textbf{CLIP text-image direction similarity} ($\text{CLIP}_{\bm{dir}}$),
(2) \textbf{CLIP image similarity} ($\text{CLIP}_{\bm{img}}$), and (3) \textbf{CLIP output similarity} ($\text{CLIP}_{\bm{out}}$). To evaluate model specificity, we utilize the following three metrics: (4) \textbf{Mean Squared Error} (\textbf{MSE}), (5) \textbf{Attack Success Rate} (\textbf{ASR}), and (6) \textbf{Error Attack Rate} (\textbf{EAR}). Where ASR and EAR are computed based on the MSE threshold, which is introduced in~\cite{VillanDiffusion}. Further details on all evaluation metrics are provided in Appendix~\ref{sec:metrics}.

We present the performance of backdoor models trained with different watermarking methods in Table~\ref{tbl:results}. All experiments are conducted with the poison rate set to 0.1. To assess the utility of the backdoor models, we compare their editing capabilities against InstructPix2Pix (Clean), which corresponds to the original model weights released by~\cite{instructpix2pix}. As shown in Table~\ref{tbl:results}, both StegaStamp and RoSteALS achieve performance comparable to the clean model, whereas VINE performs slightly worse than the other two methods. In model specificity, StegaStamp and RoSteALS attain high ASR while maintaining low EAR, indicating stronger and more precise backdoor behavior. In contrast, VINE exhibits the lowest ASR and the highest MSE and EAR, reflecting the weakest performance in terms of specificity. To provide a more intuitive understanding of the experimental results, we show qualitative outputs of all backdoor models in Figure~\ref{fig:visualization}. Ideally, when an original image is input into the backdoor model, it should produce the correctly edited output according to the given prompt. Conversely, when a watermarked image is input, the model should generate the predefined backdoor target. As shown in Figure~\ref{fig:visualization}, the outputs of the backdoor model trained with VINE frequently fail to meet these expectations.

\subsection{Analysis and Discussion}
\label{sec:exp_analysis}

\begin{figure}[t]
    \centering
    \scalebox{0.9}{
    \includegraphics[width=\linewidth]{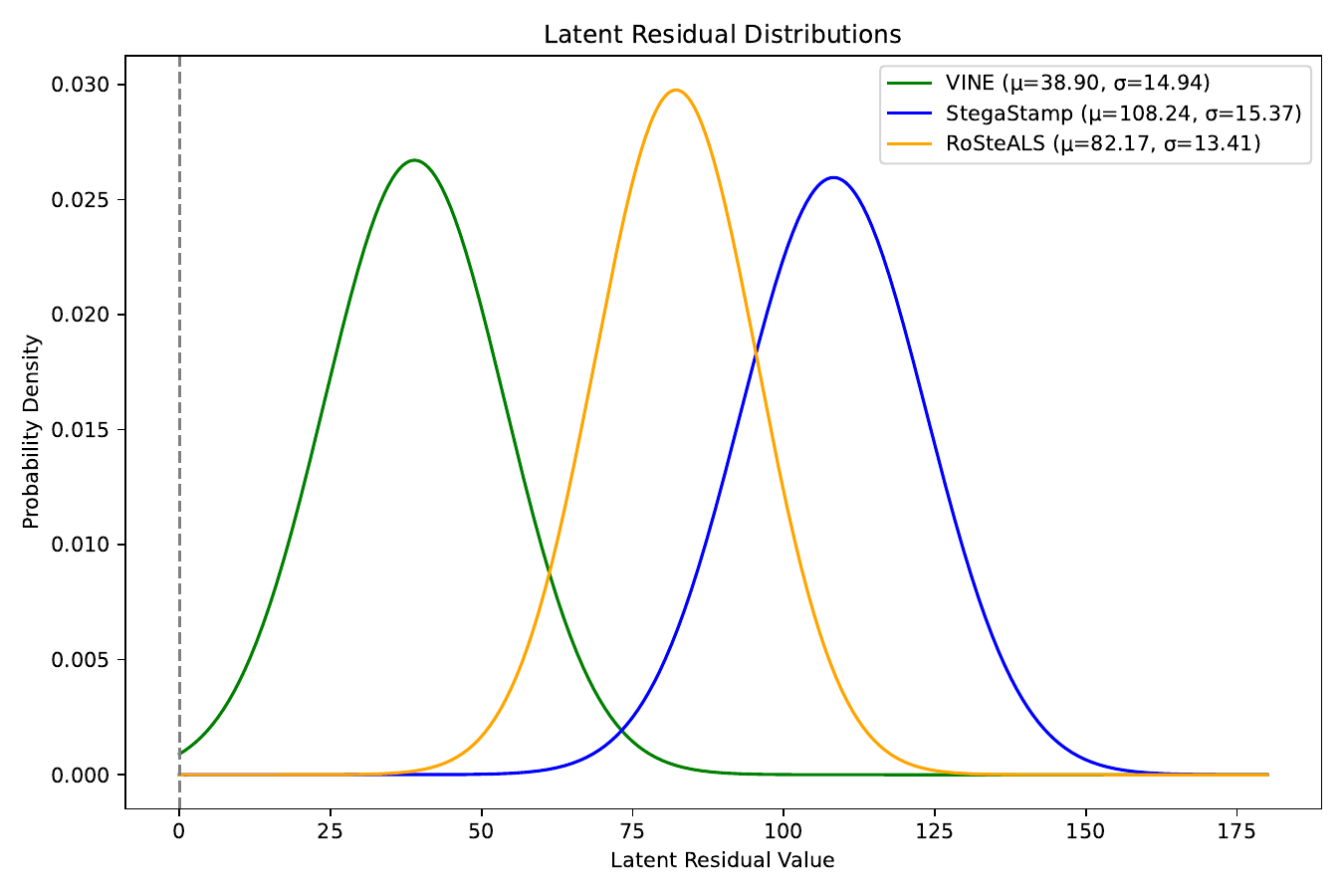}
    }
    \caption{The plot illustrates the probability density of L2 distances between original and watermarked latent representations for each watermarking method. The vertical dashed line at 0 indicates no difference.
    }
    \label{fig:distribution}
    \vspace{-10pt}
\end{figure}

We assume that the effectiveness of our attack framework primarily stems from the latent differences between the original and watermarked images. The larger the latent difference, the greater the success of the backdoor model. Let $\mathcal{E}(x)$ represent the original latent, and $\mathcal{E}(x_{w})$ denote the corresponding watermarked latent. The latent residuals for the $i$-th sample can be formulated as $r^*_i = \left\| \mathcal{E}(x_{w^*_i}) - \mathcal{E}(x_i) \right\|_2 $, where $* \in \{\text{StegaStamp}, \text{VINE}, \text{RoSteALS} \}$ represents the method used to generate watermarked image. We then collect these residuals into the set $R^* = \left\{ r^*_i \in \mathbb{R} \,\middle|\, \; i = 1, 2, \dots, N \right\}$, where $N$ represents the total data in the testing set. We compute the mean $\mu$ and standard deviation $\sigma$ of each set and plot the latent residual distribution for each watermarking model in Figure~\ref{fig:distribution}. We assume that given a sufficient number of samples, the curves can be approximated by a normal distribution ($\mathcal{N}(\mu, \sigma^2)$). According to Figure~\ref{fig:distribution}, we observe that the latent residual between original images and watermarked images embedded through StegaStamp exhibits the largest difference. This result aligns with the performance presented in Table~\ref{tbl:results}, where the backdoor model trained with StegaStamp achieves the best performance. Although a large latent distance can enhance the effectiveness of the backdoor model, the watermark should remain imperceptible in the image space. To assess the quality of the watermarked image, we present the quantitative evaluation in Table~\ref{tbl:analysis_img}. Although the quantitative values for watermarked images generated by StegaStamp and RoSteALS are relatively low, the watermarked images remain perceptually similar to their original images, as shown Figure~\ref{fig:supp_watermark} in the Appendix.

\subsection{Robustness}
\label{sec:exp_robustness}

To evaluate the robustness of our backdoor models in real-world scenarios, we assess their performance under various image distortions. Following the WAVES benchmark~\cite{WAVES}, we test three categories of distortions: (i) \textbf{Geometric}: rotation, resized-crop and erasing; (ii) \textbf{Photometric}: brightness and contrast; and (iii) \textbf{Degradation}: JPEG compression, Gaussian blur and Gaussian noise. Since our pipeline is to trigger the backdoor target when a watermarked image is provided as input, it is essential to verify that this functionality remains effective under such transformations. Due to the poor performance of the backdoor model trained with VINE, we focus on evaluating the models trained with StegaStamp and RoSteALS in our subsequent experiments. As shown in Table~\ref{tbl:robustness}, both StegaStamp and RoSteALS demonstrate strong robustness under erasing and JPEG compression, achieving high ASR and low EAR. However, StegaStamp is vulnerable to contrast changes (68.1\% ASR), while RoSteALS is susceptible to Gaussian noise (61.8\% ASR). Additionally, both methods perform poorly under rotation, resized cropping, and Gaussian blur, indicating that these distortions remain particularly challenging for our pipeline.

\begin{table}[t]
    \small 
    \centering
    \scalebox{1.0}{
    \begin{tabular}{c|ccc}
    \toprule
    Watermark & PSNR\textuparrow & SSIM\textuparrow & LPIPS\textdownarrow  \\
    \midrule
    VINE & \textbf{37.85 $\pm$ 3.15} & \textbf{0.993 $\pm$ 0.006}  & \textbf{0.003 $\pm$ 0.01}   \\
    StegaStamp & 31.88 $\pm$ 2.74 & 0.927 $\pm$ 0.040 & 0.055 $\pm$ 0.02 \\
    RoSteALS & 28.37 $\pm$ 2.78 & 0.876 $\pm$ 0.049 &  0.031 $\pm$ 0.01  \\
    \bottomrule
    \end{tabular}
    }
    \vspace{0.25em}
    \caption{Image quality comparison of watermarking models. Values for each metric are reported as mean and standard deviation.}
    \label{tbl:analysis_img}
    \vspace{-10pt}
\end{table}

\subsection{Ablation study}
\label{sec:exp_ablation}

\begin{table*}[tb]
    \centering
    \begin{tabular}{cc|c|ccc|ccc}
    \toprule
    $\bm{\mathcal{L}_{\epsilon_\theta}}$ & $\bm{\mathcal{L}_{MSE}}$ & Watermark &  $\text{CLIP}_{\bm{dir}}$\textuparrow & $\text{CLIP}_{\bm{img}}$\textuparrow & $\text{CLIP}_{\bm{out}}$\textuparrow & MSE\textdownarrow & ASR\textuparrow & EAR\textdownarrow \\
    \midrule
    \(\checkmark\) &  &  & 0.194 & 0.719 & 0.232 & 0.319  & 0.624 & 0.112 \\
     & \(\checkmark\) & StegaStamp & 0.154 & 0.684 & 0.223 & 0.042 & \textbf{0.958} & 0.008 \\
    \(\checkmark\) & \(\checkmark\) &  & \textbf{0.208} & \textbf{0.759}  & \textbf{0.255} & \textbf{0.038}  & 0.956 & \textbf{0.000} \\
    \midrule
    \(\checkmark\) &  &  & 0.164 & 0.671 & 0.194 & 0.534 & 0.342 & 0.301 \\
     & \(\checkmark\) & RoSteALS & 0.161 & 0.655  & 0.213  & \textbf{0.044}  & \textbf{0.952} & 0.024 \\
    \(\checkmark\) & \(\checkmark\) &  & \textbf{0.211} & \textbf{0.745}  & \textbf{0.252} & 0.092 & 0.894 & \textbf{0.003} \\
    \bottomrule
    \end{tabular}
    \vspace{0.5em}
    \caption{Impact of the loss functions described in Section~\ref{sec:method} under a poison rate of 0.1.}
    \label{tbl:ablation_loss}
    \vspace{-5pt}
\end{table*}

\begin{figure}[t]
    \centering
    \scalebox{1.0}{
    \includegraphics[width=\linewidth]{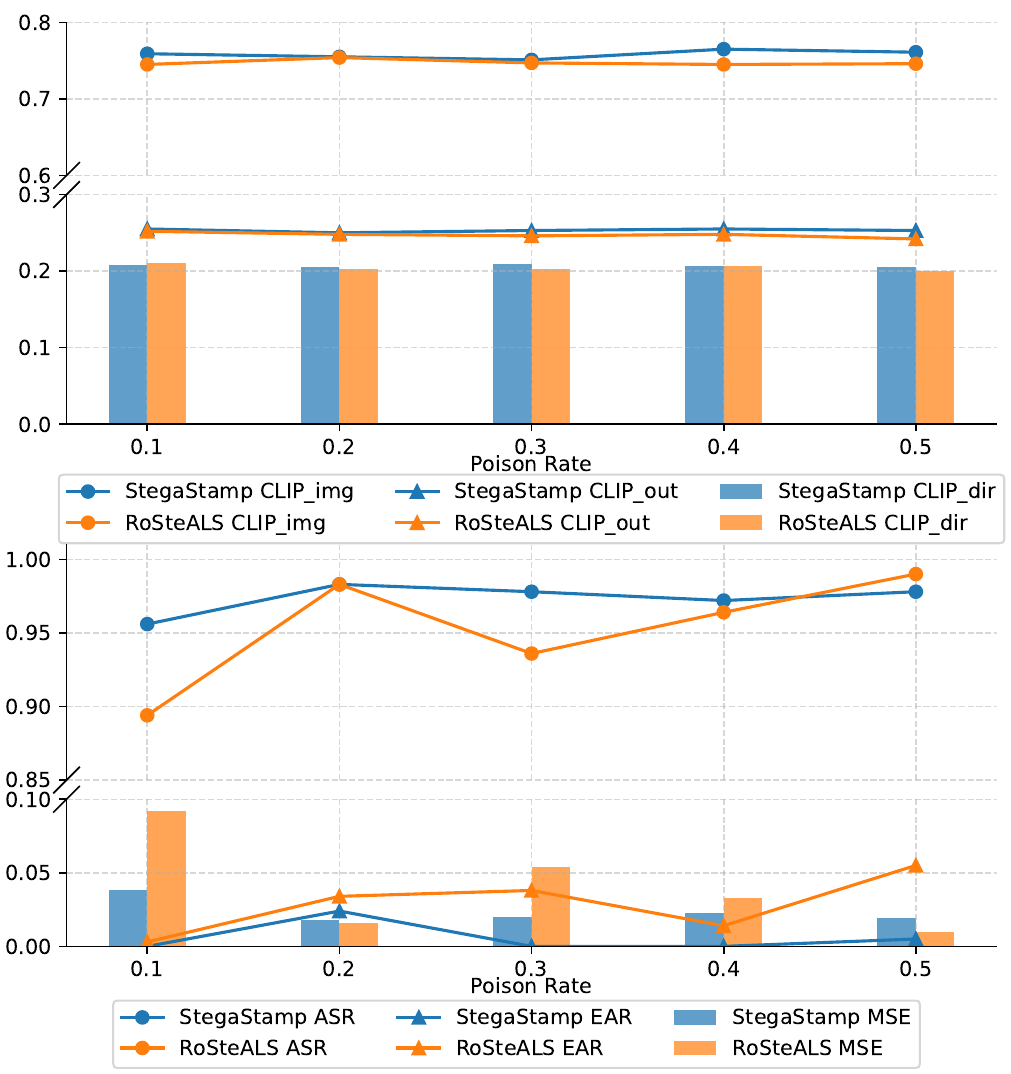}
    }
   \caption{Performance at Various Poison Rates. The top plot depicts model utility, while the bottom plot illustrates model specificity for StegaStamp (blue) and RoSteALS (orange).}
\label{ablation:poison_rate}
\vspace{-10pt}
\end{figure}

In this section, we provide the impact of different loss functions in Table~\ref{tbl:ablation_loss}, and the effect of different poison rates in Figure~\ref{ablation:poison_rate}. For the discussion of performance with multiple trigger-target pairs, please refer to Appendix~\ref{ablatio:multi-target}.

As shown in Table~\ref{tbl:ablation_loss}, both StegaStamp and RoSteALS exhibit a similar trend. When only the denoising loss $\mathcal{L}_{\epsilon_\theta}$ is applied, the attack proves largely ineffective, achieving an ASR of 62.4\% for StegaStamp and 34.2\% for RoSteALS. Although employing only the MSE loss $\mathcal{L}_{MSE}$ yields the highest ASR, it significantly compromises the model’s inherent editing capability, as indicated by a substantial drop in CLIP-based scores. In contrast, the combination of both losses enables a successful backdoor attack while preserving the model’s original editing functionality.

In Figure~\ref{ablation:poison_rate}, we present the performance of backdoor models across varying poison rates. As the poison rate increases, RoSteALS gradually closes the gap with StegaStamp in terms of ASR, achieving comparable performance at higher levels of poisoning. However, this gain is accompanied by a noticeable increase in EAR, indicating a potential risk of false positives. In contrast, StegaStamp consistently maintains a high ASR across all poison rates while exhibiting a relatively stable and lower EAR compared to RoSteALS, demonstrating greater reliability under varying poisoning conditions.

\section{Conclusion}
We present a novel framework that utilizes off-the-shelf deep watermarking models to embed invisible backdoor triggers into the image editing pipeline. Our results show that the framework not only enables effective backdoor attacks but also maintains high visual fidelity. Further analysis indicates that substantial latent residual differences are key to the attack’s success. We also evaluate the robustness of the framework under various distortions, identifying both its strengths and limitations. Future work may explore advanced watermark embedding strategies that balance strong latent space separation, minimal perceptual impact, and improved robustness.

{\small
\bibliographystyle{ieee}
\bibliography{egbib}
}

\appendix
\clearpage
\setcounter{page}{1}
\setcounter{table}{0}
\renewcommand{\thetable}{S.\arabic{table}}
\setcounter{figure}{0}    
\renewcommand{\thefigure}{S.\arabic{figure}}  
\newpage
\twocolumn[
    \centering
    \Large
    \textbf{Invisible Backdoor Triggers in Image Editing Model via Deep Watermarking}\\
    \vspace{0.5em}
    Supplementary Material \\
    \vspace{1.0em}
]

\section{Implementation Details}
\label{sec:implementation}
The secret message used for watermarking is a fixed 100-bit binary string across all watermarking models~\cite{vine, stegastamp, rosteals}. For VINE~\cite{vine} and RoSteALS~\cite{rosteals}, we utilize the officially released checkpoints. The StegaStamp~\cite{stegastamp} model, however, is trained by ourselves using the training strategy in~\cite{artificialGANfingerprint}. To ensure a fair comparison, we train StegaStamp using 100,000 images sampled from the MIRFLICKR dataset, which is the same dataset used to train RoSteALS. To fine-tune the InstructPix2Pix model~\cite{instructpix2pix}, we sample 10,000 image pairs for training and 1,000 for testing from the \texttt{timbrooks/instructpix2pix-clip-filtered} dataset. All input images are resized to $256 \times 256$ resolution. We fine-tune the pretrained InstructPix2Pix model, released by~\cite{instructpix2pix}, for 50 epochs using the AdamW optimizer with a learning rate of $1 \times 10^{-4}$. For the fine-tuning of InstructPix2Pix with different watermarking models, we set the batch size to 24 for StegaStamp and RoSteALS, and 16 for VINE, due to GPU memory constraints.

\section{Metrics}
\label{sec:metrics}
We introduce the metrics used in our experiments to evaluate model utility and specificity:
(1) \textbf{CLIP text-image direction similarity} ($\text{CLIP}_{\bm{dir}}$) measures the consistency between semantic changes in the text prompts and the corresponding visual changes in the images.
(2) \textbf{CLIP image similarity} ($\text{CLIP}_{\bm{img}}$) evaluates the similarity between edited images and their original counterparts.
(3) \textbf{CLIP output similarity} ($\text{CLIP}_{\bm{out}}$) assesses the alignment between the edited images and their associated output captions. (4) \textbf{Mean Squared Error} (\textbf{MSE}) measures the pixel-wise difference between the generated backdoor sample and the ground truth (GT) backdoor target.
(5) \textbf{Attack Success Rate} (\textbf{ASR}) denotes the percentage of watermarked images that successfully trigger the generation of the backdoor target.
(6) \textbf{Error Attack Rate} (\textbf{EAR}) evaluates the percentage of clean (non-watermarked) images that incorrectly result in the generation of the backdoor target. To evaluate ASR and EAR of the backdoor model, we adopt the MSE threshold introduced in~\cite{VillanDiffusion} to determine whether the backdoor target has been successfully generated.

Specifically, if the MSE value between the generated sample and the GT backdoor target is below the predefined threshold, the generation is considered to produce the backdoor target (assigned 1); otherwise, it is assigned 0. For example, given the threshold $\phi$, the ASR can be formulated as

\begin{equation}
\label{eq:asr}
   \text{ASR} = \frac{1}{N}\sum_{i=1}^{N}{\mathds{1}\left( \mathcal{L}_{MSE}(y', \tilde{y'_i}) < \phi \right)},
\end{equation}
where $y'$ denotes the GT backdoor target, $\tilde{y}'_i$ is the $i$-th generated backdoor target, and $N$ is the number of samples in the test set. Similarly, the EAR is formulated as
\begin{equation}
\label{eq:ear}
   \text{EAR} = \frac{1}{N}\sum_{i=1}^{N}{\mathds{1}\left( \mathcal{L}_{MSE}(y', \tilde{y_i}) < \phi \right)},
\end{equation}
where $\tilde{y}_i$ denotes the $i$-th generated edited image and $\phi$ is set to 0.1.

\begin{figure}[t]
    \centering 
    \scalebox{0.9}{
       \includegraphics[width=\linewidth]{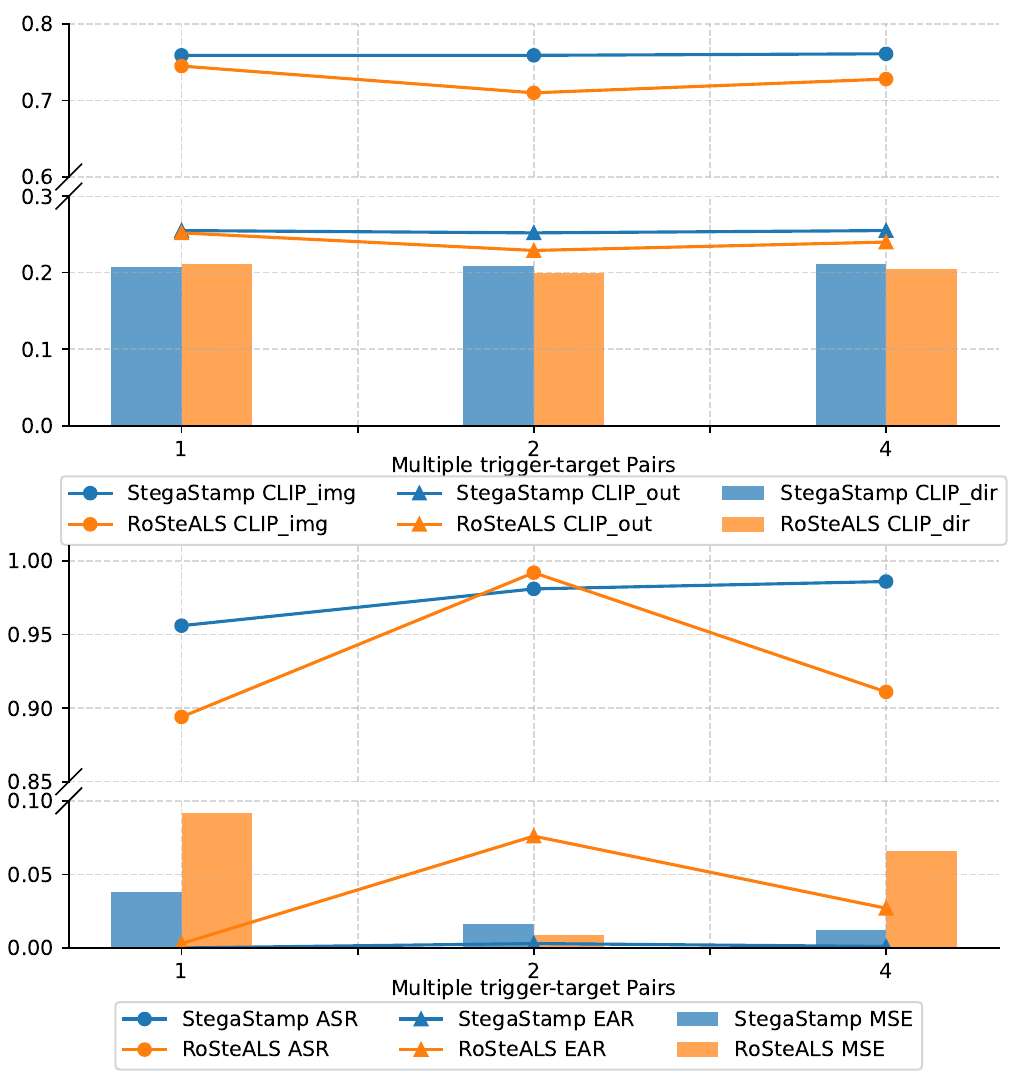}
    }
   \caption{Performance with multiple trigger-target pairs. Leveraging off-the-shelf deep watermarking models, our framework can easily extend to multiple trigger-target pairs by mapping different secret messages to distinct targets.}
\label{ablation:multi-target}
\end{figure}

\section{Multiple Trigger-Target Pairs}
\label{ablatio:multi-target}

We demonstrate that our attack framework can be readily extended to support multiple trigger-target pairs. By leveraging off-the-shelf deep watermarking models, we map distinct secret messages to corresponding target outputs. As shown in Figure~\ref{ablation:multi-target}, the ASR remains stable as the number of trigger-target pairs increases. This observation confirms that our system is capable of handling multiple embedded triggers simultaneously without compromising performance.

\begin{figure*}[tb]
    \centering
    \includegraphics[width=\textwidth]{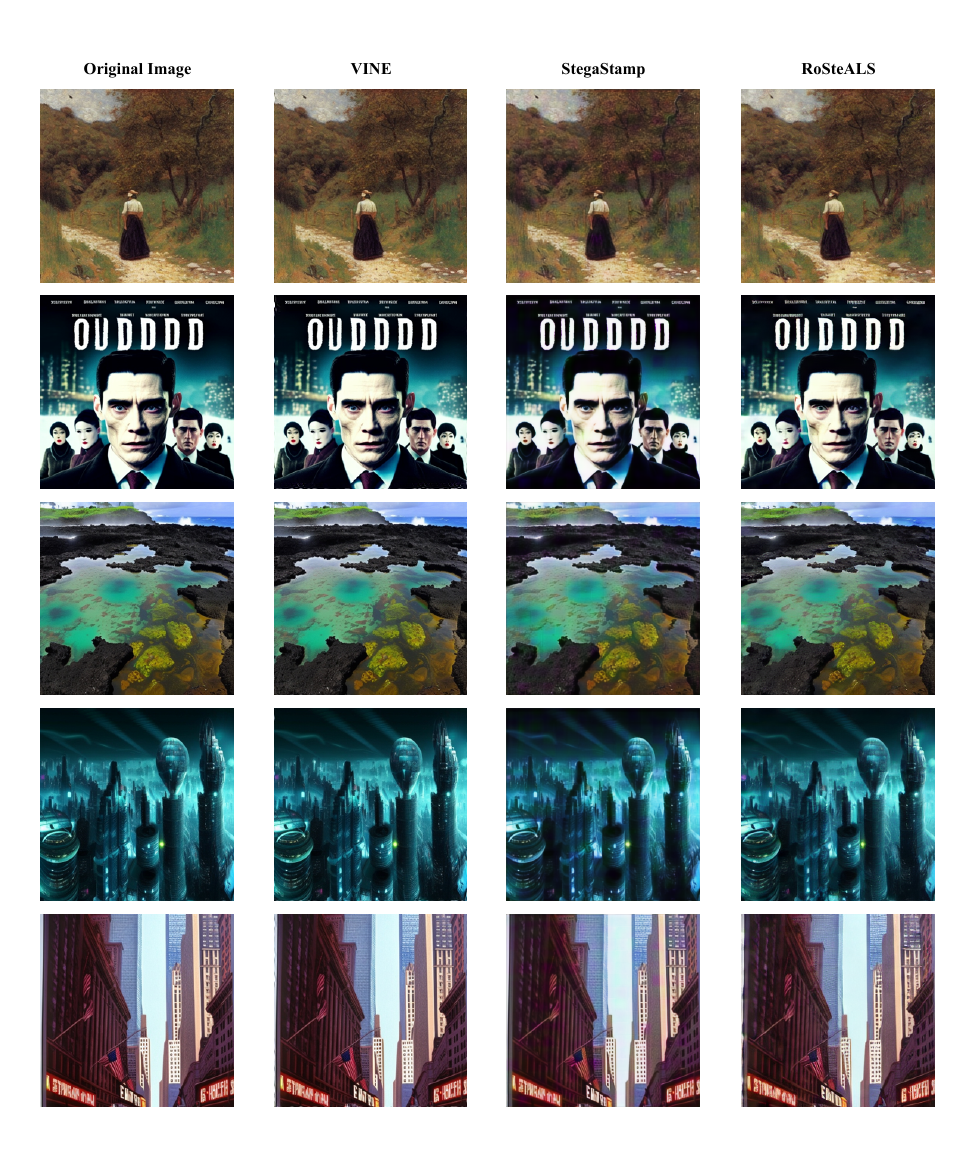}
    \caption{\textbf{Samples of Original and Watermarked Images.} Each row shows an original image and its corresponding watermarked image generated by VINE~\cite{vine}, StegaStamp~\cite{stegastamp}, and RoSteALS~\cite{rosteals}.}
    \label{fig:supp_watermark}
\end{figure*}


\end{document}